\documentclass[conference]{IEEEtran}

\usepackage{graphicx}
\usepackage{amsfonts,amssymb,amsmath,amsgen,amsopn,amsbsy,theorem,graphicx,epsfig}
\usepackage{multirow}
\usepackage{hyperref}
\usepackage{xcolor}

\ifCLASSINFOpdf

\else
\fi

\begin{document}
%
\title{In-Depth Analysis of Automated Acne Disease Recognition and Classification}
\author{\parbox{16cm}{\centering
    {\large Afsana Ahsan Jeny$^{1,2}$, Masum Shah Junayed$^{1,2}$, Md Robel Mia$^3$, Md Baharul Islam$^{1,4}$}\\
    {\normalsize
    $^1$ Department of Computer Engineering, Bahcesehir University, Istanbul, Turkey\\
    $^2$Department of CSE, University of Connecticut, Storrs, CT 06269, USA \\
    $^3$ Department of CSE, Daffodil International University, Dhaka, Bangladesh \\
    $^4$ Dept. of Computing \& Software Engineering, Florida Gulf Coast University, Fort Myers, FL 33965, USA}}
}

\maketitle

\begin{abstract}
Facial acne is a common disease, especially among adolescents, negatively affecting both physically and psychologically. Classifying acne is vital to providing the appropriate treatment. Traditional visual inspection or expert scanning is time-consuming and difficult to differentiate acne types.  This paper introduces an automated expert system for acne recognition and classification. The proposed method employs a machine learning-based technique to classify and evaluate six types of acne diseases to facilitate the diagnosis of dermatologists. The pre-processing phase includes contrast improvement, smoothing filter, and RGB to $L*a*b$ color conversion to eliminate noise and improve the classification accuracy. Then, a clustering-based segmentation method, k-means clustering, is applied for segmenting the disease-affected regions that pass through the feature extraction step. Characteristics of these disease-affected regions are extracted based on a combination of gray-level co-occurrence matrix (GLCM) and Statistical features. Finally, five different machine learning classifiers are employed to classify acne diseases. Experimental results show that the Random Forest (RF) achieves the highest accuracy of $98.50\%$, which is promising compared to the state-of-the-art methods.
\end{abstract}


%
\IEEEpeerreviewmaketitle

\section{Introduction}\label{sec1}
{Acne is a prevalent dermatological condition affecting individuals irrespective of gender. Its onset is primarily attributed to bacterial presence, obstruction of hair follicles by oil, accumulation of dead skin cells, and the overproduction of sebum. Areas rich in sebaceous glands, such as the face, forehead, chest, back, and shoulders, are commonly afflicted \cite{islam2023acne}. The heterogeneity in acne lesion morphology necessitates accurate differentiation for effective assessment and intervention. Moreover, lesion location may provide insights into underlying conditions, emphasizing the importance of precise acne identification \cite{hameed2020hybrid}.}
{According to the global acne market study \cite{GlobalAcneMarket}, over 90\% of the global population is affected by acne. In 2006, acne afflicted 612 million individuals, a figure that rose by 10\% over the subsequent decade. Projections suggest that by 2026, nearly 23 million people in India alone will suffer from acne, representing a compound annual growth rate of 0.5\%. Consequently, global treatment costs are rising. Dermatologists often diagnose specific acne conditions and enumerate acne lesions through visual inspection, a method that can be laborious and imprecise. Two leading skin analysis tools in cosmetic surgery are VISIA from Canfield \cite{visia2020skin} and ANTERA 3D from Miravex \cite{antera2020Acne}. They aid in treating skin pores, acne scars, and offer anti-aging solutions. However, these tools are expensive and demand specialized knowledge for effective utilization \cite{park2019objective}. Several diagnostic methods, such as fluorescence light-based photography \cite{lucchina1996fluorescence} and multi-spectral imaging \cite{fujii2008extraction}, strive to offer dermatologists clearer insights into acne lesions and their characteristics. Despite their precision, these methods require significant manual effort from dermatologists, especially considering the varied scales, shapes, and locations of acne lesions and the diversity in skin tones and lesion types.}

Researchers have recently proposed machine and deep learning-based techniques to solve these issues and assist dermatologists. Some of these methods are for acne severity\cite{nguyen2021severity, wen2022acne}, acne grading\cite{lin2021acne, alzahrani2022attention} and only acne detection (acne has or not) \cite{yadav2021hsv}. However, limited studies have been performed for the acne classification \cite{junayed2022transformer, hameed2020hybrid, shen2018automatic, junayed2019Acnenet, isa2021acne}. The most recent acne detection method, e.g., \cite{yadav2021hsv}, utilized a human visual system (HSV) based model for segmentation and support vector machine (SVM) and two CNN-based models for detection. Although this method's accuracy is higher than other methods, they utilized only {$120$} Acne images for training and a total of {$200$} images for the experiments. In addition, they only detect acne, which can be divided into two classes: normal skin and acne; they do not categorize the numerous kinds of acne. In another study, Hameed et al. \cite{hameed2020hybrid} used image processing techniques for segmentation and a Naive Bayes classifier for four types of Acne classification. In their experiments, they used only 40 images and obtained 93.42\% accuracy. Therefore, the main challenges are a proper acne recognition system and a large-scale dataset containing different acne types from the above observations. Besides, it is also significant to improve the performance of the clinical trial of the automated acne disease classification.

\begin{figure*}[htb]
    \centering
    \includegraphics[width=0.9\textwidth]{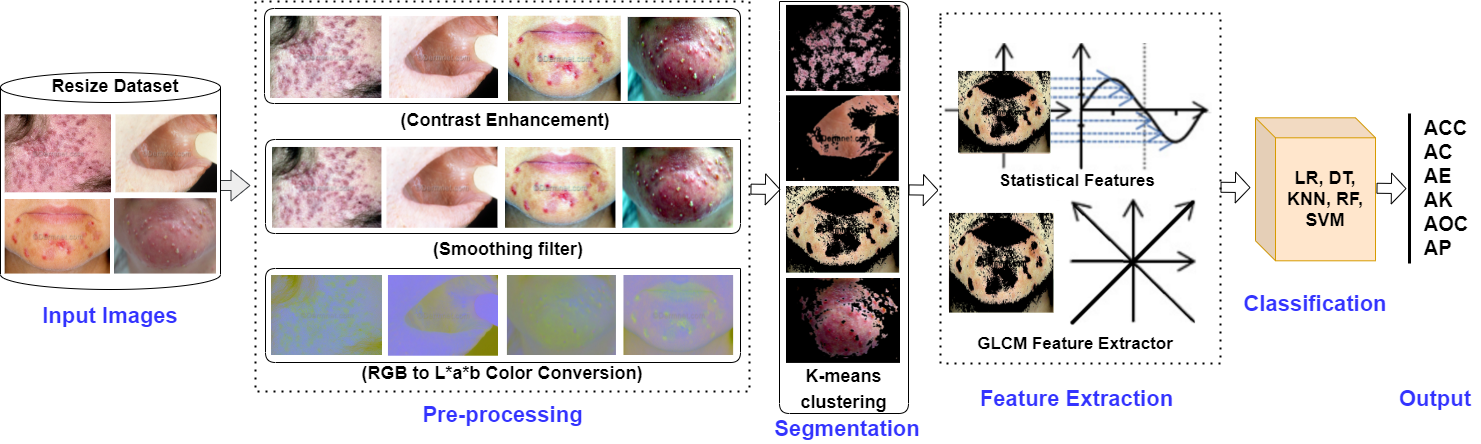}
    \caption{An overview of the proposed expert system for recognizing and classifying acne disease. Input images are pre-processed through contrast enhancement, smoothing filter, and L*a*b color conversion and then segmented using k-means clustering. The GLCM matrix and Statistical feature extraction methods are employed to extract features. The five classifiers, namely logistic regression, decision tree, K-nearest neighbors, random forest, and support vector machine, are utilized to classify acne diseases.}
    \label{fig:system}
\end{figure*}

\textbf{Contributions.} Motivated by the abovementioned challenges, we propose an automated acne disease recognition and classification method for six classes and a comprehensive analysis using a large-scale comparative dataset. Firstly, the proposed system used some pre-processing before applying K-means clustering to segment different acne classes. For feature extraction, the GLCM and statistical feature selection methods are employed. Five machine learning classifiers, including decision tree (DT), k-nearest neighbors (KNN), support vector machine (SVM), random forest (RF), and logistic regression (LR), are used to classify six acne classes and compare the acquired results. Our main contributions are given below. 
\begin{itemize}
    \item We introduce an automated expert system to recognize acne disorders through segmentation and categorize them into six different classes from facial acne images.
    \item We applied contrast enhancement, smoothing filter, and RGB to $L*a*b$ color conversion to better represent the acne regions and reduce the noise in the preprocessing stage. Then, the acne-affected areas are segmented using k-means clustering.
    \item {Two standard feature extraction methods, i.e., GLCM, and statistical features, are utilized together for acne-segmented images to improve classification accuracy in every classifier.
    \item To demonstrate the system's robustness, five machine-learning classifiers are employed to classify acne disease images with high accuracy. }
\end{itemize}

\section{Proposed Method}\label{3}
Fig. \ref{fig:system} depicts the step-by-step visualization of the proposed expert system for acne disease recognition and classification using machine learning-based techniques. {A pre-processing procedure is first performed on our collected dataset. During the pre-processing stage, the image contrast is enhanced, a smoothing filter is applied, and the color space is transferred from RGB to $L*a*b$. To avoid overfitting, we used some augmentation methods.} Next, k-means clustering is applied to segment the input images. Finally, two feature extraction methods are applied together to the segmented areas, and these features are used to classify the acne types. In the following sub-sections, every step of the proposed architecture is explained in detail. 

\begin{figure}[htb]
    \centering
    \includegraphics[width=0.4\textwidth]{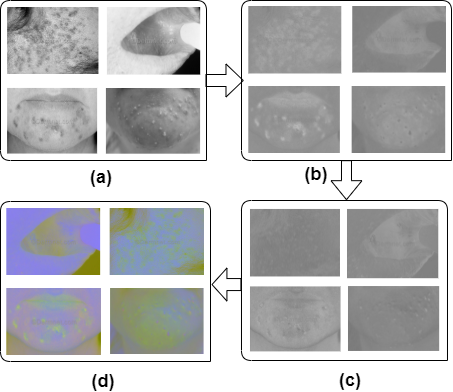}
    \caption{Visualization of the L*a*b color conversion process step by step. Here, (a), (b), (c), and (d) denote the visualization of L, a, b, and L*a*b conversions, respectively.}
    \label{fig:lab}
\end{figure}

\textbf{Pre-processing:} {In the pre-processing step, the medical images are primarily treated to decrease the distortion caused by noises and enhance the essential information in the original image. Three image processing techniques, such as contrast enhancement, smoothing filter, and RGB to $L*a*b$ color conversion, are included in our system to depict acne areas better and eliminate the noises. The guided image filtering \cite{he2012guided} is employed in our system as a contrast enhancement technique to reduce the input image noise while maintaining the borders of acne diseases by increasing contrast. Enhancing the images' edge makes it simpler for the machine learning models to learn and discriminate between various types of acne diseases. After that, the blurring and noise reduction in images are accomplished with the help of a smoothing filter \cite{cadena2017noise}. In our experiment, the Gaussian filter is applied as a smoothing filter that blurs images without removing information and eliminates noises. Some samples of contrast enhancement and smoothing filter are depicted in the pre-processing portion of Fig. \ref{fig:system}.} Then, the images are converted from RGB to $L*a*b$ color space \cite{hosen2018detection}. It can be written as per color conversion from RGB color space to $XYZ$ color space.
$
\begin{bmatrix}
 X \\ Y \\ Z
\end{bmatrix}
= 
\begin{bmatrix}
3.240479 & -1.537150 & -0.498535 \\ -0.969256 & 1.875992 & 0.041556 \\ 
0.055648 & -0:204043 & 1.057311
\end{bmatrix}
\begin{bmatrix}
R \\ G \\ B
\end{bmatrix}$
\noindent To convert $XYZ$ color space into $L*a*b$, we can assume that $X_n, Y_n$, and $Z_n$ are tri-incentive standards as per the situation. We can redraft it as the following equation \ref{eq:3}:
\begin{equation}\label{eq:3}
\begin{aligned}
f(x)=\left\{x^{\frac{1}{3}}, \text { if } x>0.008856\right.\\
7.787 x+\frac{16}{116}, \text { if } x>0.008856
\end{aligned}
\end{equation}
\noindent Then $L*a*b$ equations (\ref{eq:4}, \ref{eq:5}, \ref{eq:6}) can be written as:
\begin{equation}\label{eq:4}
\begin{gathered}
L=\left\{116\left(\frac{J}{J_{n}}\right)^{\frac{1}{3}}-16, \text { if } \frac{J}{J_{n}}>0.008856\right. \\
903.3\left(\frac{J}{J_{n}}\right), \text { if } \frac{J}{J_{n}}>0.008856
\end{gathered}
\end{equation}
\begin{equation}\label{eq:5}
    a* = 500f(\frac{I}{I_n}) - f(\frac{J}{J_n})
\end{equation}
\begin{equation}\label{eq:6}
    b* = 200f(\frac{J}{J_n}) - f(\frac{K}{K_n})
\end{equation}
\noindent the color contrast is defined by $I, J,$ and $K$, and the white colorless specular reflection is described by $I_n$, $J_n$, and $K_n$. {This conversion provides a more accurate representation of the input images and the acne disease areas, which improves the final classification accuracy dramatically. The visualization of some $L*a*b$ color conversion images are shown in Fig. \ref{fig:lab}.}

\begin{figure*}[htb]
    \centering
    \includegraphics[width=0.9\textwidth]{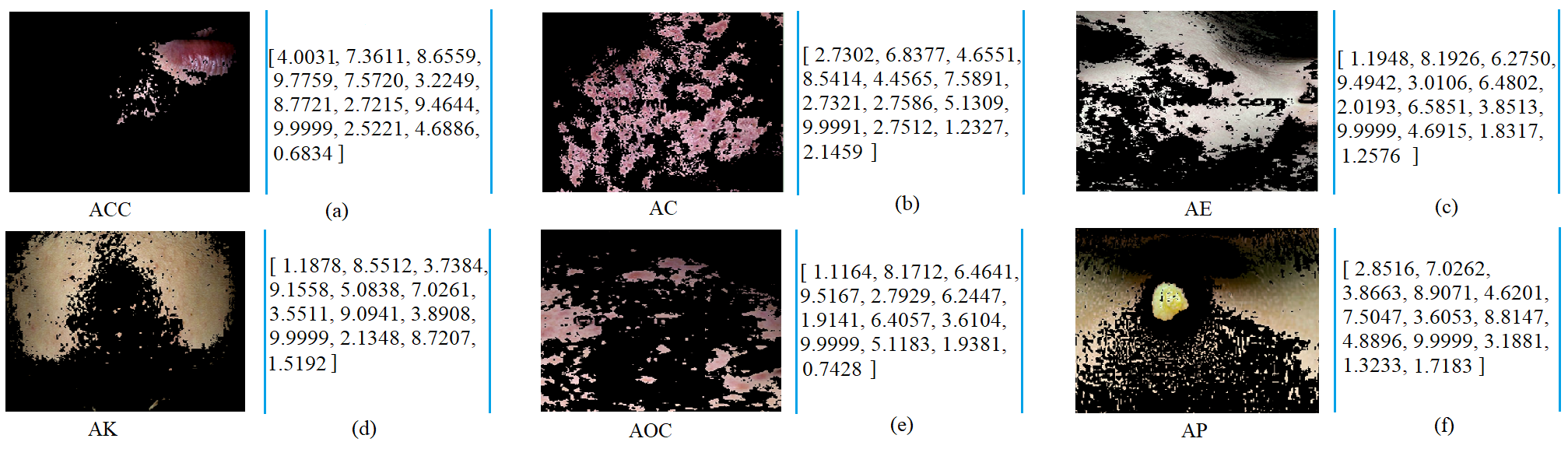}
    \caption{Segmentation and feature extraction results of acne disease are taken as examples. The following list including, (a), (b), (c), (d), (e), and (f) are extracted feature matrics ($C$, $\rho$, $E$, $S$, $H$, $\mu$, $\sigma$, $\sigma^2$, $K$, RMS, Smoothness, Skewness, and cluster shade (Cs)) of ACC, AC, AE, AK, AOC, and AP, respectively.}
    \label{fig:prep}
\end{figure*}

\vspace{-0.8em}
\subsection{Segmentation}
{In the acne image segmentation process using K-means clustering, statistical features were primarily introduced based on texture analysis. Acne lesions often exhibit distinct textural patterns compared to healthy skin. These features provided a foundation for differentiating acne from unaffected areas, optimizing the segmentation step. This method is used to segment the image into different regions so that we can do further processing.} Here, the main target is to define K centers, one for each cluster. Since the other locations cause different results, some repetitive experiments are used in this paper. K-means aims to divide $C$ observations among k clusters. Assume a set of data ${X_1,....,X_C}$ with $C$ instances (observations) of a specific D-dimensional Euclidean value of $X$. This approach clusters data by comparing the inter-point measurements of data points to the measurements between points outside the cluster. To institutionalize this idea, we consider a template, $V_k$, affiliated with the $K^{th}$ cluster. These experiments represent the clusters' boundaries. The aim is to identify the middle of each region so that square numbers from every value are kept at least to the cluster center $V_k$ nearest to it \cite{burney2014k}.
The objective function $J$ is generally referred to as distortion measure, is given as follows (\ref{eq:7}):
\begin{equation}\label{eq:7}
    J = \sum ^C_{i=1} \sum ^K_{j=1} (||X_i - V_k||)^2
\end{equation}
\noindent where $(||X_i - V_k||)^2$ is the Euclidean distance between $X_i$ and $V_k$. Also, K is the number of data points in $k$ clusters, and $C$ is the number of the cluster centers.

\subsection{Feature Extraction}
{Feature extraction is pivotal in machine learning, guiding the optimal use of variables for specific classifiers. In image classification, the Gray Level Co-occurrence Matrix (GLCM) texture attributes are frequently employed \cite{liu2021feature}. However, GLCM can exhibit reduced accuracy near class boundaries. To address this, we integrate statistical features with GLCM attributes. This combination not only enhances textural feature extraction from images but also mitigates GLCM's aforementioned limitation.} Let, $f(x, y)$ is a two-dimensional image with M$\times$N pixels and L gray levels. In $f(x, y)$, suppose $(x1, y1)$ and $(x2, y2)$ to be two pixels, the length between them is d, and the point among them and the ordinate is h. At that point turns into a GLCM $P(i, j, d, h)$:
Let us assume $f(x, y)$ is a digital image in two dimensions of $Z$ by $G$ pixels with gray level numbers $L$. More specifically, we surmise that $(x1, y1)$ and $(x2, y2)$ are two pixels in $f(x, y)$, the distance is $T$, and the point between the two and the ordinate is $O$. At that point, a GLCM $Q (i, j, T, O)$ turns according to \cite{liu2021feature} is composed as \ref{eq:8}:
\begin{equation}\label{eq:8}
\begin{split}
    Q (i, j, T, O) = (x1, y1), (x2, y2) \in Z \times G: \\ T, O, f(x1, y1) = i, f(x2, y2) = j
\end{split}
\end{equation}

\begin{table*}[htb]
\centering
\caption{Description of the five classifiers used in the proposed system.}
\scalebox{0.85}{
\begin{tabular}{|l|p{17.0cm}|}
\hline
Classifiers & Description \\ 
\hline
LR & LR is used to predict the value of the deciding y variable when the y variable is y [0, 1], the negative class is 0, and the positive class is 1. Similarly, it can also be used for multi-classification to identify the value of y when y [0,1,2,3] is provided. Our LR model is Multinomial and the logistic function is: logit(k) = lnk - ln (1-k).\\
\hline
SVM &  SVM transforms nonlinear data into a higher-dimensional space in which it is linearly separable, thus improving its classification accuracy. To divide several classes, hyperplanes are used, and the optimum hyperplane is the one that maximizes the margin between the classes. Because of its strong generalization capacity, it is utilized effectively in a wide range of classification areas. In this experiment, we have used the linear kernel function, and the parameter value of C is $55$. \\
\hline
RF & The random forest is competent in locating null values from a vast number of datasets and can give a more accurate result than other methods for data mining. The tree's maximum height is an unlimited integer, and the number of attributes is zero, which is selected randomly in this work. The size of each RF bag is equal to the percentage of training data size. \\
\hline
DT &  As with a tree, DT shapes include leaves or decision nodes that may be selected. It is composed of internal and external nodes in the system. The offspring nodes that visit the following nodes are among the internal nodes that make decisions. We have utilized the Gini index as a degree of impurity. Entropy is for information gain, the minimum number of the split is 2, and the maximum depth of the tree is infinity. Note that the expansion of nodes continues until all leaves are pure or when the minimum number of split samples is reached for all leaves. \\
\hline
KNN & As an example, when KNN predicts the class label of new input, it is compared with the similarity of new input to the input’s samples from the training set. This condition is fulfilled if the new input is identical to samples from a previously trained set. {Usually, KNN classification performance is not very excellent for the classification problem. Here, we have set the value of K as 7, and the Manhattan distance is utilized.} \\ 
\hline
\end{tabular}}
\label{tab:svmm}
\end{table*}

We utilized five GLCM features in our experiments, such as Contrast C, Correlation ($\rho$), Energy E, Entropy S, Homogeneity H, and cluster shade (Cs) Below are their corresponding equations \ref{eq:9} - \ref{eq:13}.
\begin{equation}\label{eq:9}
   \text{C} = \sum^{L-1}_{i=1} \sum^{L-1}_{j=0}(i - j)^2 P(i - j)
\end{equation}
\begin{equation}\label{eq:10}
    \rho = \frac{\sum^{L-1}_{i=1} \sum^{L-1}_{j=0}i.j.P(i,j) - \mu _x . \mu _y}{\sigma _x . \sigma _y}
\end{equation}
\begin{equation}\label{eq:11}
    \text{E} = \sum^{L-1}_{i=1} \sum^{L-1}_{j=0} (i,j)^2
\end{equation}
\begin{equation}\label{eq:12}
    \text{S} = \sum^{L-1}_{i=1} \sum^{L-1}_{j=0} P(i,j) \log P(i,j)
\end{equation}
\begin{equation}\label{eq:13}
    \text{H} = \sum^{L-1}_{i=1} \sum^{L-1}_{j=0} \frac{P(i,j)}{1 + (i - j)^2}
\end{equation}
\begin{equation}\label{eq:14}
    \text{Cs} = \sum^{L-1}_{i=1} \sum^{L-1}_{j=0} {(i + j -i.p(i,j)-j.p(i,j))}^3 p(i,j)
\end{equation}
\noindent where $\mu _x$, $\mu _y$, $\sigma _x$ and $\sigma _y$ are the sum of estimated and modification standards for the row and column matrix correspondingly.

Based on the findings in \cite{mutlag2020feature}, we have selected a few statistical features that can be used to identify Acne diseases. Here, we have provided the statistical features which are selected in our system. The related equations for the mentioned statistical features are as equations (\ref{eq:15} - \ref{eq:19}): 
\begin{equation}\label{eq:15}
   \text{Mean,} \mu = \frac{\sum^N_{i=1}GS_i}{N}
\end{equation}
\begin{equation}\label{eq:16}
    \text{Standard deviation,} \sigma = \sqrt{\frac{{\sum^N_{i=1}}(G{S_i} - \mu)^2}{N}}
\end{equation}
\begin{equation}\label{eq:17}
    \text{Variance,} \sigma^2 = \frac{{\sum^N_{i=1}}(G{S_i} - \mu)^2}{N}
\end{equation}
\begin{equation}\label{eq:18}
   \text{Kurtosis,} K = \frac{\frac{1}{N}{\sum^N_{i=1}(G{S_i} - \mu)^4}}{(\frac{1}{N}{\sum^N_{i=1}(G{S_i} - \mu)^2})^2} - 3
\end{equation}
\begin{equation}\label{eq:19}
    \text{Skewness,} \gamma = \frac{\mu - N_o}{\sigma}
\end{equation}
\noindent we have addressed the $N$ number of pixels in defective districts, where $GS$ represents the grayscale shading power of a pixel, $\mu$, and $N_o$ addressed the mean and mode of the grayscale color intensity of all pixels separately.

{Fig. \ref{fig:prep} depicts the outcomes of our selected feature extraction techniques (GLCM and Statistical) for six acne disease classes that are absolutely robust to increase the classification accuracy. Moreover, in Table \ref{table:performn}, it is observed that we were able to achieve higher accuracy in every classifier by using our combined feature extraction methods. The feature map in Fig. \ref{fig:prep} represents the value of GLCM features (Contrast C, Correlation ($\rho$), Energy E, Entropy S, Homogeneity H, and cluster shade (Cs)), Statistical features (mean $\mu$, standard deviation $\sigma$, variance $\sigma^2$, kurtosis K, and skewness $\gamma$), RMS, and smoothness respectively.}

\textbf{Classification:} {For the classification of acne diseases, five machine learning classifiers, namely LR, DT, KNN, SVM, and RF, are utilized. These classifiers are shortly explained in Table \ref{tab:svmm}.} 

\section{Experiments} \label{4}
\textbf{Dataset:} We collect $320$ images from the public platform Dermnet (\url{https://dermnet.com/}), and $120$ images from New Zealand Dermatologists (\url{https://dermnetnz.org/}) prior to applying augmentation. 
{A total of $2100$ images are used for our experiments. There are $410, 300, 340, 330, 370$, and $350$ images for ACC,  AC, AE, AK, AOC, and AP classes, respectively. The dataset is divided into two parts: training and testing. 80\% images (1680 images) are used for training and 20\% images (420 images) for testing. Utilizing a 5-fold cross-validation \cite{marcot2021optimal} technique and averaging the results, the proposed method's performance is assessed.}


\textbf{Experimental Setup:} Experiments in this paper are done using the following computing system: Intel Core i$9$ Central Processing Unit (CPU) operating at $3.60$ GHz, $64$ GB of RAM, and an NVIDIA Geforce RTX $2080$ Super GPU with $8$GB of GPU memory. All the detection and classification computations are implemented using MATLAB (R$2016$a) software.

\textbf{Training Details: } 
We divided our dataset into two parts: training and testing. We employed the holdout approach \cite{raschka2018model} to specify the number of data allocated for training and testing \cite{awwalu2019holdout}. Around $66\%$ of the sample data set ($1387$ color images) is used for training, and the rest ($713$ color images) is used for testing.
{Additionally}, the original training {dataset} is split into two smaller subsets: {one for testing, and another one for validation.} Next, about $66\%$ of the training {set ($916$ images) is} utilized for the {classification}, and the rest ($471$ images) {is used} for error evaluation. To {identify} the best classifier, holdout techniques \cite{raschka2018model} {is} applied multiple times. {After extensive experiments, the performance of all of the classifiers is analyzed.} 

\textbf{Evaluation {Metrics}:} {For evaluating the performance of each classifier, accuracy cannot be asserted as a thorough measurement for the estimation of the open exhibition. This is because, it would not be ideally suited for evaluating identification characteristics acquired from unequal class distribution datasets, because for example,} the quantities of samples in different classes vary greatly. Other assessment matrices for evaluating the output of a classifier are built on the Confusion Matrix \cite{junayed2020eczemanet}, as described in \cite{tax2002using}. For a two-class scenario, a binary Confusion Matrix (CM) shows the number of true positives (TPs), false negatives (FNs), false positives (FPs), and true negatives (TNs). {CM for a multi-class classification (W), can be written as the following equation:}
\begin{equation}
    W = [b_{ij}]_{nxn}
\end{equation}
The multi-class CM (W) is a $n \times n (n > 2)$ square matrix. It has n rows and n columns, {totally including} $n^2$ entries. There is no simple way to calculate the number of FPs ($\sum_{j=1, j\neq i}^{n} {b_{ji}}$), FNs ($ \sum_{j=1, j\neq i}^{n} {b_{ij}}$), TPs ($b_{ii}$), and TNs ($\sum_{j=1, j\neq i}^{n} \sum_{k=1, k\neq i}^{n} {b_{jk}}$) for the multi-class {CM}. The results of TPs, FNs, FPs, and TNs are determined and as per the regulations for multi-class matrix as described in {the CM}. The final {CM} dimension is $2x2$, and the mean values are kept in n confusion matrices for each class. An uncertainty matrix, also known as an error matrix, can be utilized for statistical classification. The is used to measure the accuracy ($\frac{(TP + TN)}{(TP + TN + FP + FN)} \times 100(\%)$), precision ($\frac{(TP)}{(TP + FP)} \times 100(\%)$), specificity ($\frac{(TN)}{(FP + TN)} \times 100(\%)$), sensitivity ($\frac{(TP)}{(FN + TP)} \times 100(\%)$), {false positive rate (FPR)} ($\frac{(FP)}{(FP + TN)} \times 100(\%)$), and {false negative rate (FNR)} ($\frac{(FN)}{FN + TP} \times 100(\%)$). 
\vspace{-0.5em}
\section{Results and Discussion}\label{5}
\subsection{Performance Analysis}
Table \ref{table:cm} shows the binary CM for each of the six acne disease classes of RF utilizing our dataset. As can be seen in this table, the performance of our classifiers is competitive on the acne dataset, as shown by the large number of TP values we obtained in our research and the low misclassification rates in each of the classes.

\begin{table}[htb]
\centering
\caption{The binary Confusion Matrix (CM) for every class.}
\scalebox{0.9}{
\begin{tabular}{|c|c|c|c|c|c|c|c|c|c|}
\hline
\textbf{Class} & \multicolumn{4}{c|}{\textbf{Matrix}} & \textbf{Class} & \multicolumn{4}{c|}{\textbf{Matrix}} \\ \hline
\multirow{4}{*}{ACC} & \multirow{4}{*}{Actual} & \multicolumn{3}{c|}{Predicted} & \multirow{4}{*}{AC} & \multirow{4}{*}{Actual} & \multicolumn{3}{c|}{Predicted} \\ \cline{3-5} \cline{8-10} 
 &  &  & + & - &  &  &  & + & - \\
 &  & + & 77 & 5 &  &  & + & 57 & 3 \\
 &  & - & 3 & 335 &  &  & - & 3 & 357 \\ \hline
\multirow{4}{*}{AE} & \multirow{4}{*}{Actual} & \multicolumn{3}{c|}{Predicted} & \multirow{4}{*}{AK} & \multirow{4}{*}{Actual} & \multicolumn{3}{c|}{Predicted} \\ \cline{3-5} \cline{8-10} 
 &  &  & + & - &  &  &  & + & - \\
 &  & + & 63 & 5 &  &  & + & 64 & 2 \\
 &  & - & 4 & 348 &  &  & - & 2 & 352 \\ \hline
\multirow{4}{*}{AOC} & \multirow{4}{*}{Actual} & \multicolumn{3}{c|}{Predicted} & \multirow{4}{*}{AP} & \multirow{4}{*}{Actual} & \multicolumn{3}{c|}{Predicted} \\ \cline{3-5} \cline{8-10} 
 &  &  & + & - &  &  &  & + & - \\
 &  & + & 72 & 2 &  &  & + & 68 & 5 \\
 &  & - & 2 & 344 &  &  & - & 2 & 345 \\ 
\hline
\end{tabular}}
\label{table:cm}
\end{table}


\begin{figure}[htb]
    \centering
    \includegraphics[width=0.4\textwidth]{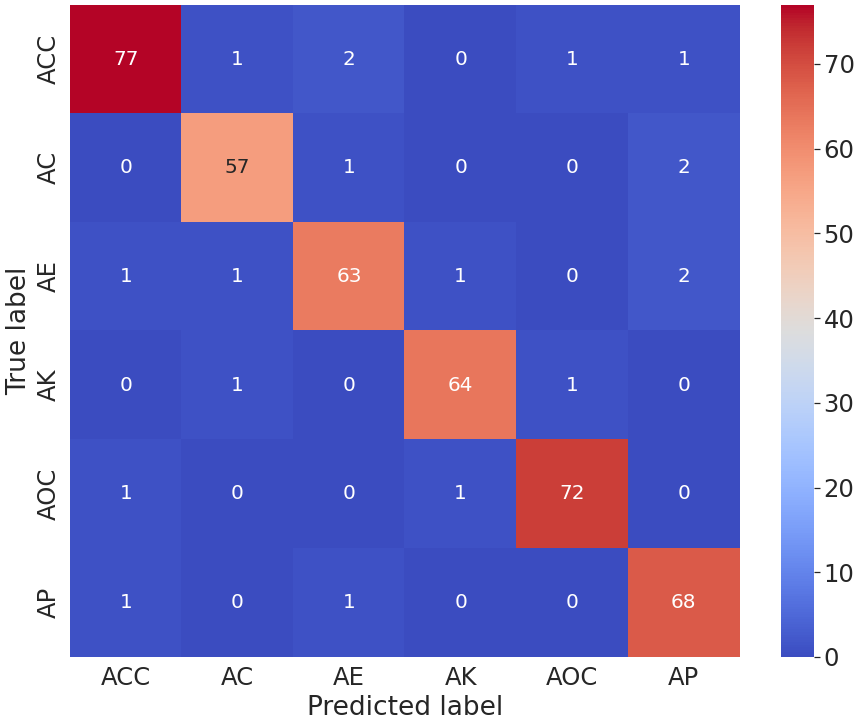}
    \caption{Multi-class confusion matrix {for the RF classifier}.}
    \label{fig:multicm}
\end{figure}

\begin{table}[htb]
\centering
\caption{Classifier's performance on different acne classes in terms of Accuracy, Precision, Sensitivity, Specificity, FPR, and FNR.}
\scalebox{0.65}{
\begin{tabular}{|c|c|c|c|c|c|c|c|}
\hline
\textbf{Classifiers} & \textbf{Classes} & \textbf{Accuracy (\%)} & \textbf{Precision (\%)} & \textbf{Sensitivity} & \textbf{Specificity} & \textbf{FPR (\%)} & \textbf{FNR (\%)} \\ \hline
\multirow{6}{*}{LR} & ACC & 97.14 & 89.39 & 89.39 & 98.35 & 1.65 & 10.61 \\
 & AC & 97.34 & 94.31 & 95.08 & 98.09 & 1.91 & 4.92 \\
 & AE & 95.91 & 84.62 & 89.19 & 97.11 & 2.89 & 10.81 \\
 & AK & 96.73 & 85.48 & 88.33 & 97.90 & 2.10 & 11.67 \\
 & AOC & 98.16 & 95.40 & 94.32 & 99.00 & 1.00 & 5.68 \\
 & AP & 95.91 & 86.96 & 90.91 & 97.01 & 2.99 & 9.09 \\ \hline
\multicolumn{2}{|c|}{Total Average} & 96.87 & 89.36 & 91.20 & 97.01 & 2.99 & 8.80 \\ \hline
\multirow{6}{*}{DT} & ACC & 96.31 & 84.85 & 87.50 & 97.64 & 2.36 & 12.50 \\
 & AC & 96.73 & 93.50 & 93.50 & 97.81 & 2.19 & 6.50 \\
 & AE & 96.32 & 85.90 & 90.54 & 97.35 & 2.65 & 9.46 \\
 & AK & 98.36 & 93.55 & 93.55 & 99.06 & 0.94 & 6.45 \\
 & AOC & 97.75 & 94.25 & 93.18 & 98.75 & 1.25 & 6.82 \\
 & AP & 96.52 & 89.13 & 92.13 & 97.50 & 2.50 & 7.87 \\ \hline
\multicolumn{2}{|c|}{Total Average} & 96.99 & 90.20 & 91.73 & 98.02 & 1.92 & 8.27 \\ \hline
\multirow{6}{*}{KNN} & ACC & 96.72 & 86.36 & 89.06 & 97.88 & 2.12 & 10.94 \\
 & AC & 96.11 & 91.87 & 92.62 & 97.28 & 2.72 & 7.38 \\
 & AE & 97.55 & 89.74 & 94.59 & 98.07 & 1.93 & 5.41 \\
 & AK & 97.75 & 88.71 & 93.22 & 98.37 & 1.63 & 6.78 \\
 & AOC & 98.16 & 94.25 & 95.35 & 98.76 & 1.40 & 4.65 \\
 & AP & 97.55 & 92.39 & 94.44 & 98.25 & 1.75 & 5.56 \\ \hline
\multicolumn{2}{|c|}{Total Average} & 97.22 & 90.55 & 93.21 & 98.10 & 1.90 & 6.79 \\ \hline
\multirow{6}{*}{SVM} & ACC & 98.77 & 93.94 & 96.88 & 99.06 & 0.94 & 3.12 \\
 & AC & 97.96 & 95.93 & 95.93 & 98.63 & 1.37 & 4.07 \\
 & AE & 96.93 & 89.74 & 90.91 & 98.06 & 1.94 & 9.09 \\
 & AK & 98.36 & 91.94 & 95.00 & 98.83 & 1.17 & 5.00 \\
 & AOC & 98.57 & 96.55 & 95.45 & 99.25 & 0.75 & 4.75 \\
 & AP & 97.75 & 93.48 & 94.51 & 98.49 & 1.51 & 5.49 \\ \hline
\multicolumn{2}{|c|}{Total Average} & 98.06 & 93.59 & 94.78 & 98.72 & 1.28 & 5.22 \\ \hline
\multirow{6}{*}{RF} & ACC & 98.13 & 93.91 & 96.26 & 98.53 & 1.47 & 3.74 \\
 & AC & 98.57 & 95.02 & 95.00 & 99.17 & 0.83 & 5.00 \\
 & AE & 97.86 & 92.65 & 94.03 & 98.58 & 1.42 & 5.97 \\
 & AK & 99.05 & 96.97 & 96.97 & 99.44 & 0.56 & 3.03 \\
 & AOC & 99.07 & 97.30 & 97.31 & 99.42 & 0.58 & 2.69 \\
 & AP & 98.33 & 93.15 & 97.14 & 98.57 & 1.43 & 2.86 \\ \hline
\multicolumn{2}{|c|}{Total Average} & \textbf{98.50} & \textbf{94.83} & \textbf{96.12} & \textbf{98.95} & \textbf{1.04} & \textbf{3.88} \\ \hline
\end{tabular}}
\label{table:performn}
\end{table}

Fig. \ref{fig:multicm} represents the multiclass CM for the RF classifier where X and Y axes represent the predicted and true levels, respectively. The misclassification rate for the RF classifier is limited. The number of correctly recognized images in ACC and AOC classes is $77$ and $72$, respectively, while the misclassification value is only $5$ and $2$, respectively. Similarly, the misclassification values for AC, AE, AK, and AP are just $3, 5, 2$, and $2$ correspondingly, demonstrating the RF’s competitive performance.

Table \ref{table:performn} displays the performance of five machine learning models, namely, DT, KNN, SVM, RF, and LR in terms of accuracy, precision, sensitivity, specificity, FPR, and FNR for each of the Acne types (i.e., ACC, AC, AE, AK, AOC, and AP). The average accuracy of LR, DT, KNN, SVM, and RF are $96.87\%, 96.99\%, 97.22\%, 98.06\%$, and $98.50\%$ respectively, according to the data presented in this table, which again RF shows the best performance. RF exceeds not only in terms of accuracy but also in other evaluation metrics, such as precision ($94.83\%$), sensitivity ($96.12\%$), specificity ($98.95\%$), FPR ($1.04\%$), and FNR ($3.88\%$), respectively. This Table also includes the individual accuracy for each of the diseases. For example, AK got the highest accuracy of $99.07\%$ while AE, has the lowest accuracy, with $97.86\%$ for RF classifier. 

\begin{figure}[htb]
    \centering
    \includegraphics[width=0.48\textwidth]{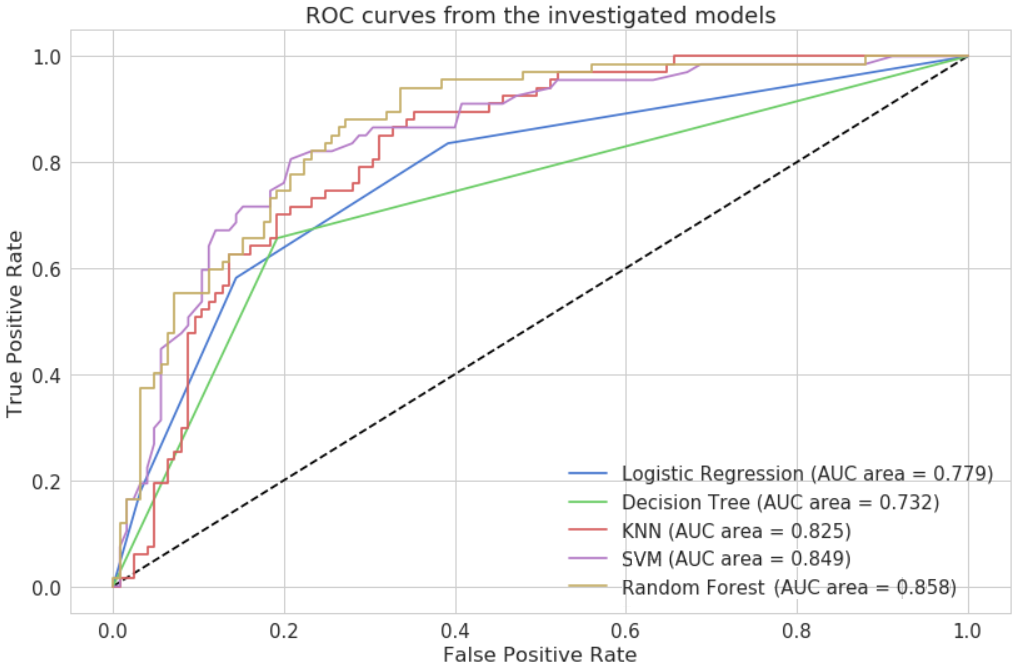}
    \caption{AUC-ROC curves of five different ML classifiers.}
    \label{fig:multirc}
\end{figure}

To further analyze the results of our method, we utilized the receiver operating characteristic (ROC) curves \cite{feng2019decision} and calculated the area under it, known as AUC. Determining which classifier is superior on average can be estimated using the area under the ROC curve. A ROC curve depicts the connection between the FPR and the TPR at various thresholds where TPR refers to Y and FPR to X axes. Fig. \ref{fig:multirc} depicts the ROC curves and area under curves of all five classifiers used in this study. This Fig. shows that RF acquires more AUC than the other four classifiers, with $85.8\%$. On the other hand, the LR has the lowest AUC ($77.9$). Furthermore, the AUC for KNN is $2.4\%$ lower than that of the SVM ($84.9$ vs. $82.5$) and $9.3\%$ higher than that of the decision tree ($82.5$ vs. $73.2$).

\begin{figure*}[htb]
    \centering
    \includegraphics[width=0.79\textwidth]{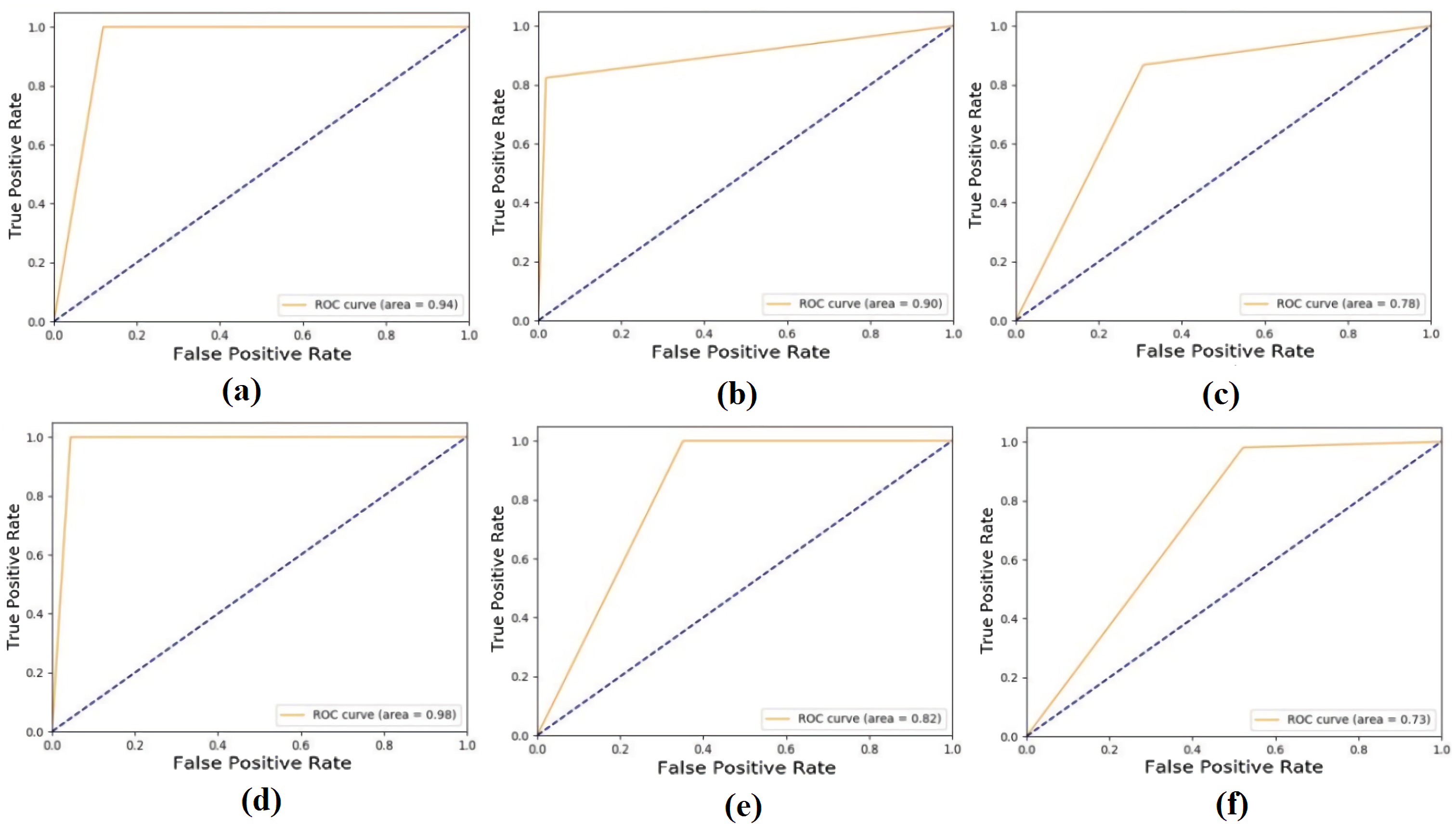}
    \caption{The AUC-ROC curves for six acne classes, (a)-(f): the ROC curves of ACC, AC, AE, AK, AOC, and AP using RF classifier.}
    \label{fig:rc}
\end{figure*}

In Fig. \ref{fig:rc}, the RF's performance was further validated across all Acne disease classes by the ROC curve, which helped to make our claim in contribution more strong. Based on our observations, we discovered that the AUC of every class is excellent; among them, the Acne Keloidalis class has the most significant amount of AUC ($98\%$), and the Acne Pustular class has the lowest amount of AUC ($73\%$). Finally, the AUC for the RF class as a whole is $85.8\%$ on average.

\begin{table}[htb]
\centering
\caption{The performance of the feature extraction methods on different classifiers (LR, DT, KNN, SVM, and RF).}
\scalebox{0.9}{
\begin{tabular}{|c|c|c|c|c|c|}
\hline
Feature Extraction & LR & DT & KNN & SVM & RF \\ \hline
Statistical & 87.56\% & 88.76\% & 89.93\% & 90.02\% & 90.61\% \\ \hline
GLCM & 92.13\% & 93.31\% & 94.67\% & 95.11\% & 95.31\% \\ \hline
Statistical+GLCM & \textbf{96.87\%} & \textbf{96.99\%} & \textbf{97.22\%} & \textbf{98.06\%} & \textbf{98.50\%} \\ \hline
\end{tabular}}
\label{tab:fmethods}
\end{table}

\subsection{{Qualitative Analysis of Feature Extraction}}
{Table \ref{tab:fmethods} shows the performance analysis utilizing feature extraction techniques. When the Statistical and GLCM feature approaches are used individually, the system's performance is worse than when the two methods are used together, as shown in Table \ref{tab:fmethods}. The GLCM approach fails to recognize image boundaries, whereas the Statistical method performs well when used for texture classification and edge detection. Consequently, their individual performance is lower, but their combined performance is remarkable in every classifier. For instance, when Statistical and GLCM methods perform separately in RF, the accuracy is 90.61\% and 95.31\%,  respectively. On the other hand, during their combination performance, the accuracy is raised by roughly 8\% (98.5\% vs. 90.61\%) and 3\% (98.5\% vs. 95.31\%), respectively.}

\begin{figure}[htb]
    \centering
    \includegraphics[width=0.45\textwidth]{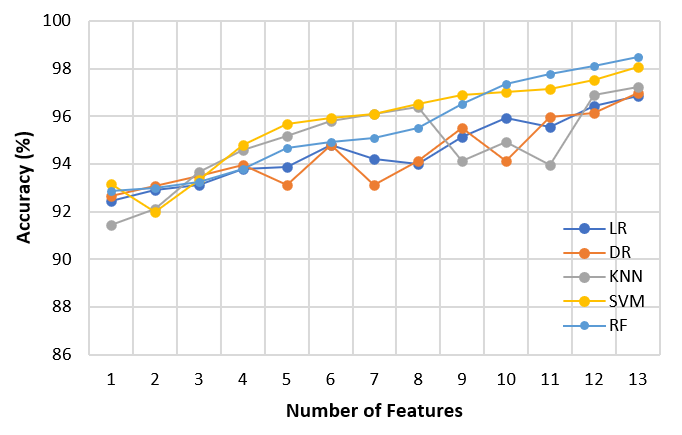}
    \caption{The average accuracy of the images using Statistical and GLCM features together in different classifiers (LR, DT, KNN, SVM, and RF).}
    \label{fig:gstatistical}
\end{figure}

{Fig. \ref{fig:gstatistical} portrays the performance investigation of accuracy vs. the number of features we employed. RF is an ensemble model, so it is projected to perform excellently. From Fig. \ref{fig:gstatistical}, we can see that as the number of features rises, the performance of the RF model does not degrade. Similarly, SVM also performs well on our dataset without showing any fluctuation. On the other hand, in LR, DR, and KNN, during feature analysis, they showed some fluctuations though their accuracy is good. It is most likely due to a violation of the feature's independence. However, it is worth mentioning that utilizing a high number of features may result in a worse performance of the model on the test dataset.}

\subsection{Comparison}
{The previous studies on acne detection and classification are depicted with their employed images in Table \ref{table:co}. In 2016, Abas et al. \cite{abas2016acne}, and Hameed et al.\cite{hameed2020hybrid} used machine learning classifiers, and their obtained accuracy was 85.5\% and 93.42\%, respectively. Shen et al.\cite{shen2018automatic}, Junayed et al.\cite{junayed2019Acnenet}, and Isa et al.\cite{isa2021acne} employed deep learning classifiers and their obtained accuracy were 91.95\% (VGG), and 91.35\% (Proposed CNN), 95.89\%, and 91.25\%, respectively. However, it is important to mention that these strategies were developed using private datasets that are not publicly accessible. As a result, this comparison is expressed from different datasets. We attempted to contact these researchers to get the implementation parameters but was not reachable.}

\begin{table}[htb]
\centering
\caption{Comparison with state-of-the-art on different datasets.}
\scalebox{0.9}{
\begin{tabular}{|c|c|c|c|}
\hline
\textbf{Approaches} & \textbf{Year} & \textbf{Dataset size} & \textbf{Accuracy} \\ \hline
Abas et al.\cite{abas2016acne} & 2016 &  \begin{tabular}[c]{@{}c@{}}17 acne images, \\ 6 classes\end{tabular} & 85.5\% \\ \hline
Shen et al.\cite{shen2018automatic} & 2018 &  \begin{tabular}[c]{@{}c@{}}3000 skin-\\ nonskin images\end{tabular} & \begin{tabular}[c]{@{}c@{}}91.95\% (VGG)\\ 91.35\% ( CNN)\end{tabular} \\ \hline
Junayed et al.\cite{junayed2019Acnenet} & 2019 &  \begin{tabular}[c]{@{}c@{}}360 images, \\ 5 classes\end{tabular} & 95.89\% \\ \hline
Hameed et al.\cite{hameed2020hybrid} & 2020 & \begin{tabular}[c]{@{}c@{}}40 images, \\ 4 classes\end{tabular} & 93.42\% \\ \hline
Isa et al.\cite{isa2021acne} & 2021 &  \begin{tabular}[c]{@{}c@{}}215 images, \\ 4 classes\end{tabular} & 91.25\% \\ \hline
\textbf{Ours (Proposed)} & 2022 &  \textbf{\begin{tabular}[c]{@{}c@{}}440 images, \\ 6 classes (Original)\end{tabular}} & \textbf{98.50\% (RF)} \\ \hline
\end{tabular}}
\label{table:co}
\end{table}


\begin{table}[htb]
\centering
\caption{Comparison with state-of-the-art on same dataset.}
\scalebox{0.95}{
\begin{tabular}{|l|l|l|l|l|}
\hline
\textbf{Datasets} & \textbf{Classes} & \textbf{Size} & \textbf{Approaches} & \textbf{Accuracy} \\ \hline
\multirow{2}{*}{Acne Classes} & \multirow{2}{*}{5} & \multirow{2}{*}{1800} & AcneNet \cite{junayed2019Acnenet} & 95.89\% \\
 &  &  & Our System & \textbf{97.13\%} \\ \hline
\multirow{2}{*}{Ours} & \multirow{2}{*}{6} & \multirow{2}{*}{2100 } & AcneNet \cite{junayed2019Acnenet} & 96.79\% \\
 &  &  & Our System & \textbf{98.50\%} \\ \hline
\end{tabular}}
\label{table:comparison}
\end{table}

To have a fair comparison between our method with state-of-the-art, we must implement state-of-the-art methods on the given dataset. After searching deeply, we got one source code, and one dataset from this related work \cite{junayed2019Acnenet}. Therefore, we only compared the proposed method with the AcneNet\cite{junayed2019Acnenet} and used the same dataset for implementation. Moreover, we utilized their source code on their dataset using our method. Table \ref{table:comparison} represents the evaluated result of the comparison. It can be seen that our system, through RF, is applied on the AcneNet dataset with 1800 images and five different Acne disease classes achieved 1.24\% more accuracy than AcneNet \ (97.13\% vs. 95.89\%). On the other hand, while AcneNet is applied to our dataset, our system also showed higher performance in terms of accuracy (98.50\%) with 2100 images and six different Acne disease classes.


\subsection{Misclassification}

\begin{figure}[htb]
    \centering
    \includegraphics[width=0.49\textwidth]{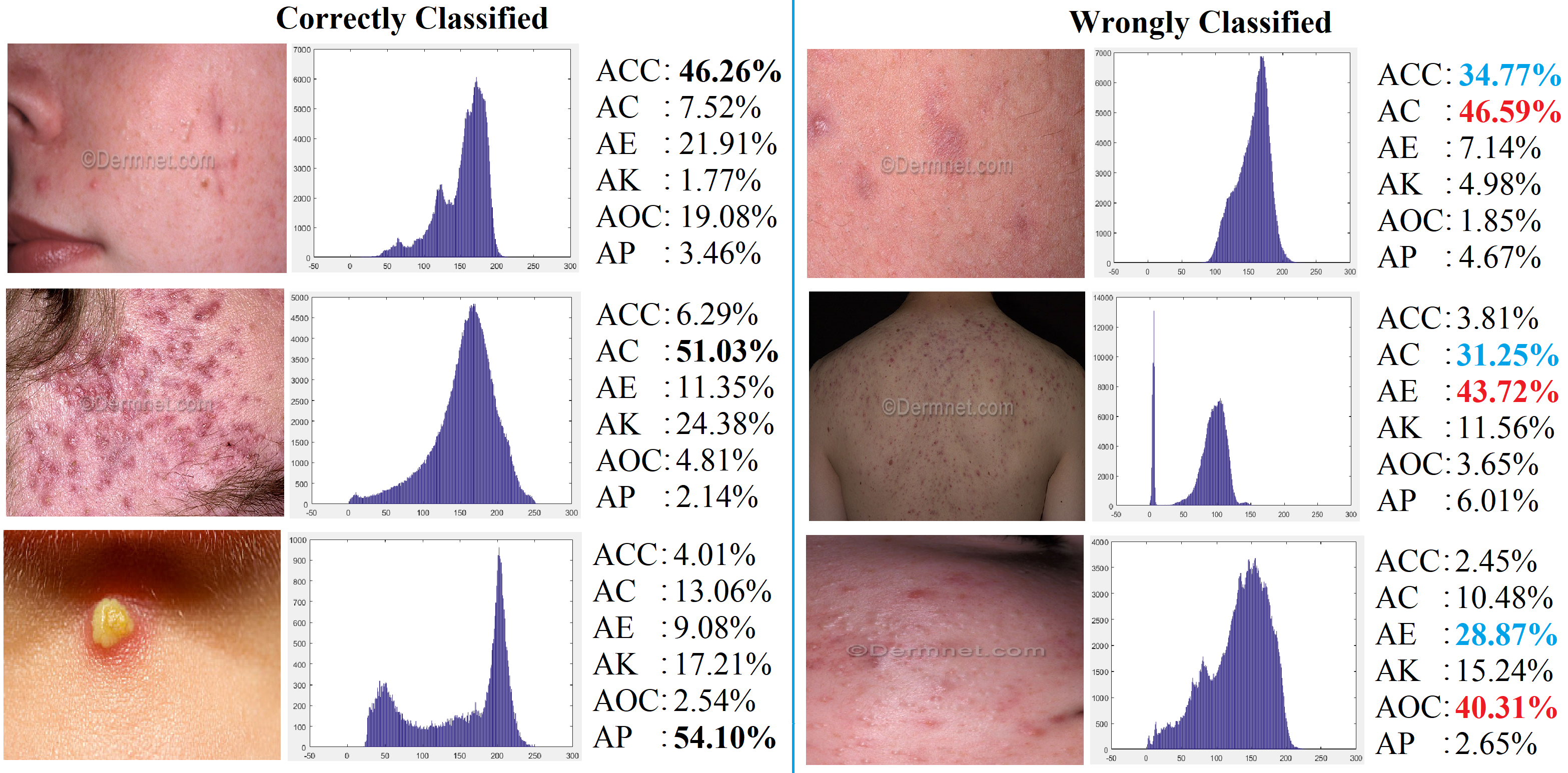}
    \caption{The classification and misclassification results of the proposed system. We demonstrated 6 acne images. Col 1-3: correctly classified images, their histograms, and scores (bold scores for the correct classes). Col 4-6: wrongly classified images, their histograms, and scores (red scores for the misclassified classes and blue scores for the correct classes) respectively.}
    \label{fig:copare}
\end{figure}

Fig. \ref{fig:copare} depicts the classification and misclassification performance of our proposed model with the histogram on the acne disease images. Six acne disease images have been taken into consideration, where three of them are predicted correctly (first column), and the rest failed in prediction (fourth column). Still, our system does not identify some of the images with severe and blurry acne. For example, the fourth column's first, second, and last images are a little bit blurry and have severe acne. For this reason, these images are misclassified into AC, AE, and AOC classes, respectively, instead of ACC, AC, and AE classes. However, they are wrongly classified, but our findings are not much different from the scores of the actual images ($46.59$\% vs. $34.77$\%, $43.72$\% vs. $31.25$\%, and $40.31$\% vs. $28.87$\%). The severity of acne is readily apparent from the histogram. For example, the first image (col 1) has less acne on its left and right, so we got values between $50$ to $200$.

\section{Conclusion}\label{6}
This article presented an automated system that recognizes and classifies six acne diseases. {A pre-processing step that included contrast enhancement, smoothing filter, and $L*a*b$ color conversion was performed to remove noise from the input images and provide better visualization.} We extracted the GLCM and statistical features before performing segmentation using k-mean clustering. Finally, extracted features were used for training five classifiers to recognize and classify acne diseases. The RF classifier achieved $98.50\%$ accuracy compared to other classifiers, a promising performance. However, our proposed system is unable to distinguish specific acne disease images. Therefore, possible future works can be focused on reducing the misclassification results and making a more uniform dataset with different acne classes to develop a more effective detection, recognition, and grading system for different types of acne disease.

\bibliographystyle{IEEEtran}
{\small
\bibliography{refs}}

\end{document}